\documentclass[10pt, a4paper]{article}

\usepackage{lrec-coling2024}
\usepackage{graphicx}
\usepackage{multirow}

\title{DGoT: Dynamic Graph of Thoughts for Scientific Abstract Generation}

\name{Xinyu Ning, Yutong Zhao, Yitong Liu\sthanks{\ \ Corresponding Author}, Hongwen Yang} 

\address{School of Information and Communication Engineering \\
         Beijing University of Posts and Telecommunications, China \\
         \{nxybupt, zhaoyutong, liuyitong, yanghong\}@bupt.edu.cn\\}

\abstract{
The method of training language models based on domain datasets has obtained significant achievements in the task of generating scientific paper abstracts. However, such models face problems of generalization and expensive training costs. The use of large language models (LLMs) to solve the task of generating paper abstracts saves the cost of model training. However, due to the hallucination problem of LLM, it is often necessary to improve the reliability of the results through multi-round query prompt approach such as Graph of Thoughts (GoT), which also brings additional reasoning costs. In this paper, we propose a Dynamic Graph of Thought (DGoT). It not only inherits the advantages of the existing GoT prompt approach, but also dynamically adjust the graph structure according to data characteristics while reducing model reasoning cost. Experimental results show that our method's cost-effectiveness in abstract generation tasks is only 43.7\% to 56.4\% of other multi-round query prompt approaches. Our code is available at \href{https://github.com/JayceNing/DGoT}{https://github.com/JayceNing/DGoT}.
 \\ \newline  }

\begin{document}

\maketitleabstract

\section{Introduction}

The abstract, as the essence of academic literature, aims to provide the core ideas, methods, and results of research. The specific domain concepts and professional terminology involved in scientific papers pose significant challenges for researchers who wish to generate article abstracts through automation. Nevertheless, the advancement of natural language processing (NLP) technology has made it possible to accurately summarize articles and generate abstracts through artificial intelligence (AI).


In pursuit of this target, previous methods usually involve collecting domain data and training corresponding models to complete the task of text summarization. The methods for generating scientific article abstracts can be divided into two categories: not using citation information and using citation information. For the former situation, \citet{Cohan2018ADA} uses an RNN-based encoder-decoder structural model to train abstract generation models specifically tailored to two large-scale scientific paper datasets. Similarly, \citet{Xiao2019ExtractiveSO} use LSTM-based models to fulfill abstract generation tasks for scientific papers. However, some researchers contend that incorporating citation information from articles can yield superior summary outcomes. \citet{Yasunaga2019ScisummNetAL} integrated the original paper and its citations, leveraging GCN and LSTM for crafting paper abstracts. The citation graph-based model has also been adopted by other researchers \cite{An2021EnhancingSP, Luo2023CitationSumCG}, they respectively used LSTM and Transformer modules to form an encoding and decoding model. Although their work has made significant progress in the abstract generation of scientific papers, training models for specific datasets often face generalization problems and also incur high training costs.

In recent years, large language models (LLMs) have attracted the attention of many researchers, such as GPT-3/4 \cite{Brown2020LanguageMA, Bubeck2023SparksOA}, LLaMA \cite{Touvron2023LLaMAOA}, ChatGLM \cite{Du2021GLMGL}, or InternLM \cite{2023internlm}. The large language model exhibits strong generalization ability in many natural language scenarios, which also brings new solutions to the task of generating paper abstracts. Relying on an autoregressive token-based mechanism by LLM, few-shot or zero-shot prompt engineering methods are used to solve the target problem. Many prompt approaches have been proposed to optimize the output results of LLM. Chain of Thought (CoT) \cite{Wei2022ChainOT} improves the model's ability to handle complex situations by adding problem-solving reasoning processes to prompt words. Tree of Thoughts (ToT) \cite{Yao2023TreeOT} models the reasoning process of LLM using trees, enabling the model to generate outcomes through multiple pathways and selecting the most favorable ones via evaluators. Graph of Thoughts (GoT) \cite{Besta2023GraphOT}  integrates the modeling of the above reasoning process using graphs, aggregating the advantages of different paths on the basis of trees. However, existing GoT approaches pre-define the number of edges, which means that the number of conversations initiated with LLM is predetermined, possibly resulting in needless resource consumption. To address this, the adaptive adjustment of the graph structure is imperative to cater to specific task requirements.


In this paper, We propose a Dynamic Graph of Thought (DGoT) prompt method to improve the quality of LLM-based generated literature abstracts while minimizing the cost of using the model. Concretely, we divide the abstract generation process based on LLM prompt approach into training process and reasoning process. (See section ~\ref{sec:method} for details). Compared to the fixed graph structure prompt method, our dynamic GoT has two main advantages: Firstly, our approach inherits the advantages of GoT structure and improves the reasoning performance of the model through effective prompt strategies. Secondly, our approach evaluates the specified task during the training process and dynamically determines the graph structure during the reasoning process, improving the cost-effectiveness of the LLM model.


In addition, we propose different threshold settings, including simple mean threshold and Gumbel threshold, to meet different requirements for the output of GoT. Lower thresholds tend to degenerate the graph into a single path during the reasoning process, while higher thresholds tend to maintain the original structure of the graph.


We evaluated the proposed method on the PubMedCite \cite{Luo2023CitationSumCG} dataset for scientific literature abstract generation. Compared to other prompt approaches based on multi-round query, our method achieved a two-fold improvement in the cost-effectiveness metric. In addition, we verified the consistency between the threshold setting and the resulting score, which can provide a reference for users to choose the threshold when using DGoT-based programs.


Our contributions can be summarized as follows:


\begin{itemize}
\item We propose a Dynamic Graph of Thought (DGoT) method that improves the performance of scientific literature abstract generation tasks while minimizing the cost of large language models.
\item We defined a threshold setting method to guide the generation process of dynamic graphs.
\item The experimental results on the PubMedCite dataset show that our method is more cost-effective than other multi-round prompt approaches.
\end{itemize}

\section{Background \& Notation}
\label{sec:append-how-prod}

In this section, we introduce the impact of probability and specific prompt approaches on the reasoning process of LLM. Additionally, we provide the definition of symbols used in our discussion. 

\subsection{Probability in LLM}
Following the established notation \cite{Yao2023TreeOT}, we use $\textbf{X}, \textbf{Y}, \textbf{Z}, ...$ to denote input sentence. Each sentence is consist of $n$ tokens, i.e. $\textbf{X} = [x_1, x_2, ... , x_n]$. The language model obtains the probability of the $i_{th}$ token through the previous $i-1$ tokens, and samples according to the probability to obtain the $i_{th}$ token:

\begin{equation}
    x_i \sim p_\theta(x_i|x_1, x_2, ... , x_i-1)
    \label{equ_lm}
\end{equation}

Where $\theta$ denotes the language model parameters, and $p_\theta$ represents the probability of the model predicting the $i_{th}$ token. 
The process iterates until the model's output corresponds to the stop token \cite{martins20sparse}.

The probability distribution over target tokens $p_\theta$ is calculated by softmax function using the output vector $v_i$ of the model \cite{Radford2018ImprovingLU}, i.e. Transformer \cite{Vaswani2017AttentionIA}. 
Using temperature $T$ to adjust the probability distribution \cite{Holtzman2020TheCC}:

\begin{equation}
    p_\theta(x_i|x_{1:i-1}) = \mathrm{softmax}(v_i)
    = \frac{exp(v_i/T)}{\sum_{i^{'}} exp(v^{'}_i/T)}
    \label{equ_softmax}
\end{equation}

Here, $T$ is confined to the interval $[0, 1]$, and the larger $T$ is, the stronger the randomness of the model output results, which lays the foundation for the prompt approaches that obtain better results through multiple rounds of questioning later.

\subsection{Prompt approaches}
\label{sec:promptA}

We formalize the existing prompt methods as the baseline for comparison:

\begin{equation}
    \textbf{Y} \sim P_\theta(\textbf{Y}|f_i(\textbf{X},\mathrm{prompt_i}))
    \label{equ_prompt}
\end{equation}

Where $P_\theta$ denotes the language model, $\textbf{X}$ is the basic information, and $\textbf{Y}$ is the model's output. $\mathrm{prompt}$ signifies the text content of the prompt word, which is combined with $\textbf{X}$ using function $f$ to form the input sequence.

\subsubsection{Input-Output (IO)}

Perform one round of query directly to the LLM to obtain the output $\textbf{Y} \sim P_\theta(\textbf{Y}|f_{IO}(\textbf{X},\mathrm{prompt_{IO}}))$.

\subsubsection{Chain-of-Thought (CoT)}

By refining the reasoning process, better output than IO can be obtained \cite{Wei2022ChainOT}. One implementation solution is to guide LLM in step-by-step reasoning in prompt text of $\textbf{Y} \sim P_\theta(\textbf{Y}|f_{CoT}(\textbf{X},\mathrm{prompt_{CoT}}))$.

\subsubsection{Tree-of-Thought (ToT)}

ToT explicitly decomposes the reasoning process for specific problems and models the reasoning process of the model as a tree structure \cite{Yao2023TreeOT}. Each node in this tree encapsulates a state $s = [x, z_{1,..., i}]$, which includes the input $x$ and output $z_i$ up to that point. It employs a generator $G(p_\theta,s,k)$ to initiate multiple rounds of questioning on LLM to generate $k$ output nodes. The output results of each step of the model are obtained by using CoT's prompt sampling $z_i \sim P_\theta(z_i|f_{CoT}([x,z_{1,...,i-1}],\mathrm{prompt_{CoT}}))$. These output results are then evaluated using an evaluator $\varepsilon(p_\theta, S)$ and the tree structure is expanded through BFS or DFS methods for better results.

\subsubsection{Graph-of-Thought (GoT)}
Compared to tree structures, GoT further expands the universality of the reasoning process by modeling it as a graph \cite{Besta2023GraphOT}. GoT is characterized by a tuple $(G, \mathcal{T}, \varepsilon, \mathcal{R})$ to represent its core components. $G = (V, E)$ represents the reasoning process, where $V$ corresponds to nodes, akin to the states $s$ in ToT, and $E$ signifies edges, denoting the relationships between different nodes. $\mathcal{T}(G, p_\theta)$ encompasses transformations, including aggregation, refinement, and generation. $\varepsilon(v, G, p_\theta)$ refers to an evaluator, tasked with scoring a single thought $v$. $\mathcal{R}(G, p_\theta, h)$ is responsible for ranking the top $h$ thoughts. Notably, this differs from ToT as GoT introduces the possibility of aggregating the results from different paths, thereby integrating the strengths of various thoughts. This enhancement boosts the efficiency of prompt approaches.

GoT integrates the previous prompt approaches and can be degraded to other methods under certain conditions. The existing GoT-based graph design methods fix the graph structure before reasoning. For specific tasks, if the graph structure is not complex enough, it may not achieve the expected effect, while overly complex graphs may bring additional overhead. Therefore, this method mainly aims to design a dynamic GoT for abstract generate tasks, which reduces reasoning costs as much as possible while using graph structures to enhance effectiveness.

\section{Method}
\label{sec:method}

\begin{figure*}[ht]
\begin{center}
\includegraphics[scale=1]{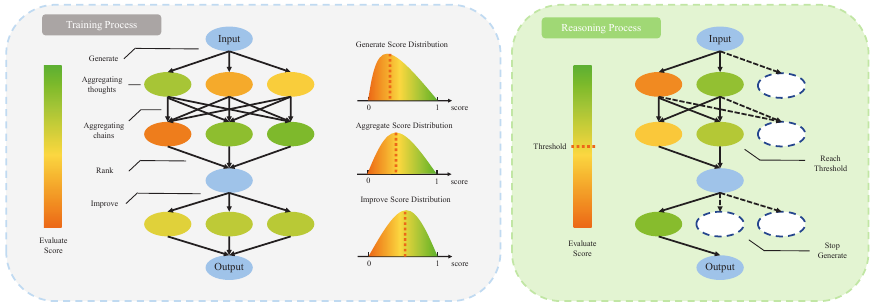} 
\caption{The overall process of our method, including the training process and reasoning process.}
\label{fig:overview}
\end{center}
\end{figure*}

\subsection{Procedure Overview}

Human cognitive researchers believe that there are two modes of people's participation in decision-making --- fast, automatic, unconscious mode (System 1), and slow, thoughtful, and conscious mode (System 2) \cite{Sloman1996TheEC}. This happens to be the same way of thinking as the current approach based on large language models, where unconscious text is output through LLM and thoughtful results are obtained through prompt approaches.

Humans develop specific thinking patterns tailored to particular tasks to enhance their problem-solving efficiency. The formation of human thinking depends on their environment and the knowledge they have acquired. Therefore, for the structure of the mind map in the prompt approaches, it is necessary to train on a particular task to obtain the optimal map structure.

Fig. \ref{fig:overview} shows the overall framework of our dynamic GoT, including the training process and reasoning process. In the training stage, we fix the GoT structure, that is, pre-define the number of nodes and edge connections in $G(V, E)$. Utilizing the training data as input for the graph, we evaluate the scores of each node in the process through $\varepsilon (p_\theta, S)$. Based on the different transformations $\mathcal{T}(G,p_\theta)$ used, we divide nodes into three categories: generating nodes, aggregating nodes, and improving nodes, and record the scores of these three types of nodes separately. Ultimately, we derive score distribution plots for the three transformations throughout the training process. We then compute the statistical characteristics of these distributions to establish a threshold.

During the inference stage, we initiate inquiries to LLM (transformations) in sequence according to the graph structure. For three different types of transformations, if the score of a query is greater than the threshold obtained during the training process, the transformation is terminated.

\subsection{Training Process}
\label{sec:train}

For three different types of transformations $\mathcal{T}$, we design the process and text of the prompt separately. Consider the example of generation transformations, which can be represented as:
\begin{equation}
    \mathcal{T}(G,p_\theta) = \textbf{Y} \sim P_\theta(\textbf{Y}|f_{Gen}(\textbf{X},\mathrm{prompt_{Gen}}))
    \label{equ_gen_prompt}
\end{equation}
Here, $\textbf{X} = [x_1,x_2,...,x_n]$ represent basic information of the article. Following SSN dataset format \citet{An2021EnhancingSP}, we define basic information of original and reference articles, which is shown in Table~\ref{basicinfo}. 

\begin{table}[b]
\begin{center}
\begin{tabularx}{\columnwidth}{l|X}

    \hline
    Category & Basic Information \\
    \hline
    \multirow{4}{*}{Original Article} & Title \\
    & Abstract \\
    & Introduction \\
    & Other Section \\
    \hline
    \multirow{2}{*}{Reference} & Title \\
    & Abstract \\
    \hline

\end{tabularx}
\caption{Basic information of the article}
 \label{basicinfo}
 \end{center}
\end{table}

\begin{figure*}[ht]
\begin{center}
\includegraphics[scale=1.3]{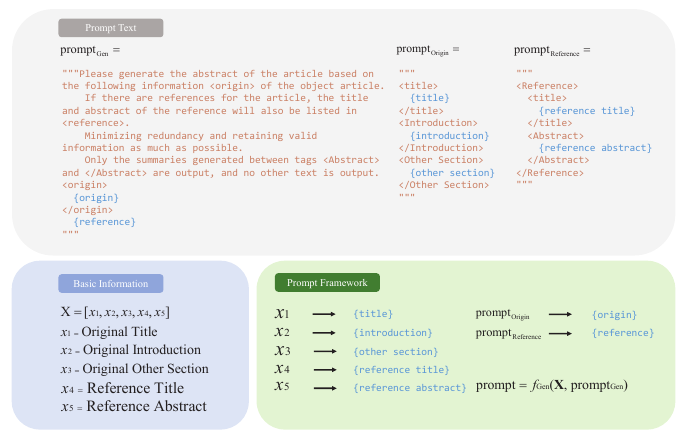} 
\caption{An example of a prompt framework.}
\label{fig:prompt}
\end{center}
\end{figure*}

$\mathrm{prompt_{Gen}}$ is the prompt text of generation transformations, and $f_{Gen}$ is the prompt framework. An example is shown in Fig. \ref{fig:prompt} (See Appendix~\ref{sec:OtherPromptFrameworks} for other prompt frameworks).

$G=(V, E)$ contains the number of times $k$ that the LLM is queried for each transformation process. The generation transformation, aggregation transformation, and improving transformation are arranged in order to form the basic structure of GoT. 

For each node $V$ in $G$, it includes the result $v$ returned by LLM under one transformation. Referring to \citet{Luo2023CitationSumCG}, we use ROUGE  \cite{Lin2003AutomaticEO} as $\varepsilon (p_\theta,S)$ to score generated abstract $v$. Due to the fact that the true abstract should be unknown to the model during the reasoning process, the ROUGE score is calculated using the generated abstract and the introduction of the original article. Therefore, during the training process, we also need to record the ROUGE scores in the generated summary and original article introduction sections of the training data. We obtain the score distribution plots of generation transformation, aggregation transformation, and improving transformation, and calculate their statistical characteristics as thresholds. For specific statistical parameters, we define them as follows.

\subsubsection{Simple Mean Threshold}

Simply use the mean of three transformation scores as the threshold.

\begin{equation}
    \mathrm{Thresh}_{\mathrm{Simple}} = \mu_{\mathrm{socre}}
    \label{mu}
\end{equation}

\subsubsection{Gumbel Threshold}
\label{subsec:gumbelthreshold}

The graph-based prompt method obtains the highest scoring result through k-round questioning. According to generalized extreme value (GEV) distribution \cite{Resnick1987ExtremeVR}, When the number of samples is large enough, Gumbel distribution can be used to model the distribution of maximum sampling values. Its PDF is:

\begin{equation}
    p(x)=\frac{1}{\beta}e^{-(z+e^{-z})}, \ \ 
    where \ \ z=\frac{x-\mu}{\beta}
    \label{GumbelPDF}
\end{equation}

During the training phase, the maximum value of the results obtained from each transformation is recorded to estimate its distribution. We calculate its mean $\mu_{Max}$ and variance $\sigma^2$, maximum likelihood estimation for $\mu$ and $\beta$ in the above equation.

\begin{equation}
    \beta^2 = \frac{6\sigma^2}{\pi^2}
    \label{beta}
\end{equation}

\begin{equation}
    \mu = \mu_{max} - \gamma \beta
    \label{mu}
\end{equation}

Where $\gamma \approx 0.577215$ is Euler-Mascheroni constant.

The CDF of Gumbel distribution is:

\begin{equation}
    F(x;\mu,\beta)=e^{-e^{-(x-\mu)/\beta}}
    \label{GumbelCDF}
\end{equation}

In the reasoning process, we define the Gumbel threshold as assuming that the maximum value has been reached with a confidence level of $p_{\mathrm{Thresh}}$. That is, $p_{\mathrm{Thresh}}=e^{-e^{-(x-\mu)/\beta}}$. The corresponding score threshold can be solved as:

\begin{equation}
    \mathrm{Thresh}_{\mathrm{Gumbel}} = \mu - \beta \mathrm{ln}(-\mathrm{ln} p_{\mathrm{Thresh}})
    \label{GumbelThresh}
\end{equation}

\subsection{Reasoning Process}

In the reasoning process, we initialize the graph structure first. We have extended the functionality of the original GoT by adding Dynamic Generate and Dynamic Aggregate modules. 

\textbf{Dynamic Generate module} directly scores the results after each inquiry with LLM, represented by $\mathcal{T}_{\mathrm{DG}}(G, p_\theta, H)$, where $H=[H_{\mathrm{Simple}}, H_\mathrm{Gumbel}]$ represents the function used to map the threshold. Corresponding to the two threshold setting methods in section~\ref{sec:train}. $H_\mathrm{Gumbel}(p)$ receives the confidence probability $p_\mathrm{Thresh}$ of the input parameter to calculate the gumbel threshold. The generation transformation repeatedly queries the LLM before the score reaches the set threshold, until the number of queries reaches $k$. 

\textbf{Dynamic Aggregate module} dynamically determines whether aggregation transformation is needed and the number of times the transformation is executed, denoted by $\mathcal{T}_{\mathrm{DA}}(G, p_\theta, H)$. Due to threshold settings, the previous step of this module may only retain one idea $v$. In that case, this step is skipped directly. The improving transformation is the same as the generation transformation, denoted by $\mathcal{T}_{\mathrm{DI}}(G, p_\theta, H)$, except for the prompt text used.

\textbf{Ranking module} is retained from the original modules of GoT, denoted by $\mathcal{R}(G, p_\theta, h)$, to preserve the best $h$ results after the generating module.

After the initialization of the graph structure, reasoning modules will be performed in order to obtain the final result.

\section{Experiments}

\subsection{Setup}
\subsubsection{Datasets}

To evaluate the effectiveness of our method, we conduct experiments on PubMedCite datasets \cite{Luo2023CitationSumCG}. Due to the license of the dataset, we did not obtain the original dataset of the author but downloaded the corresponding paper through the official API of PubMed\footnote{\href{https://pubmed.ncbi.nlm.nih.gov/download/}{https://pubmed.ncbi.nlm.nih.gov/download/}}. We organized the data according to the format in Table~\ref{basicinfo}. Finally, 10,000 original literature and reference training data pairs were obtained under the Inductive setting in PubMedCite, as well as 5224 testing data pairs.

For the test dataset, the mean ROUGE scores of the abstract and introduction parts of the source article are shown in the table~\ref{tab:AbstIntroCompare}. This ROUGE score can serve as a reference for the highest achievable score when comparing the generated results with the original paper introduction (In fact, the summarization ability of LLM is higher than this score).

\begin{table}
\centering
\begin{tabular}{llll}
\hline
\textbf{ } & \textbf{R-1} & \textbf{R-2} & \textbf{R-L}\\
\hline
Score & 0.332 & 0.132 & 0.164 \\
\hline
\end{tabular}
\caption{Comparison of ROUGE scores between the original abstract and introduction sections of the test dataset.}
\label{tab:AbstIntroCompare}
\end{table}

\subsubsection{Baselines}
\label{sec:Baselines}

We compare our method with other prompt methods, including IO, CoT, ToT, and GoT. Based on the fairness principle, the tree and graph methods will initiate the same number of LLM queries. Specifically, we set the branching factor $k=3$ and the number of levels $L=3$ for ToT. The graph method arranges 3 different types of transformations --- the generation transformation, aggregation transformation, and improving transformation in sequence, each with $E=3$ edges.

\subsubsection{Implementation Details}
\label{sec:ImplementationDetails}

Our method is implemented by Python and Pytorch. Our baseline references the design patterns of the corresponding prompt in GoT\footnote{\href{https://github.com/spcl/graph-of-thoughts}{https://github.com/spcl/graph-of-thoughts}}. The specific design of DGoT is extended based on the GoT code. We use ChatGLM2 \cite{Du2021GLMGL} as LLM for inference, and to our knowledge, ChatGLM does not use PubMedCite as training data. We deployed the ChatGLM2-6B\footnote{\href{https://github.com/THUDM/ChatGLM2-6B}{https://github.com/THUDM/ChatGLM2-6B}} model on three 24G 3090 GPUs and four 32G v100 GPUs respectively for local test. During the training and reasoning process, we did not update the model parameters but only deployed the model for inference, so the graphics memory would become a bottleneck in the input length. Therefore, we limit the input length to 20000 tokens, and any excess will be truncated. Both the top $p$ and temperature $\mathrm{T}$ of the model are fixed at 0.7.

\subsection{Training Data Distribution}

\begin{figure}[t]
\begin{center}
\hspace{-0.6cm} 
\includegraphics[scale=0.46]{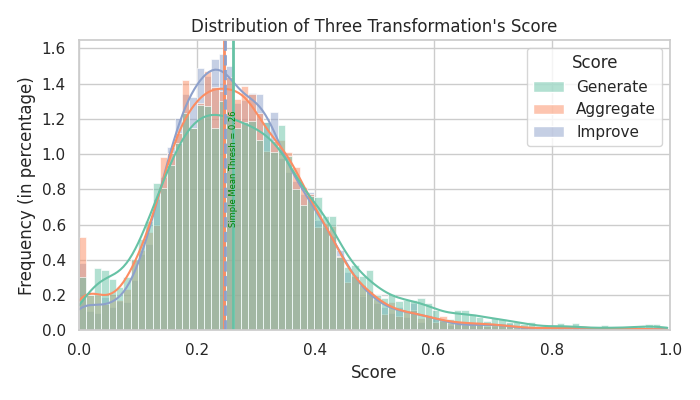} 
\caption{Score distribution of training data. The dashed line represents the mean of the corresponding transformation score.}
\label{fig3}
\end{center}
\end{figure}

\begin{figure}[t]
\begin{center}
\hspace{-0.6cm} 
\includegraphics[scale=0.46]{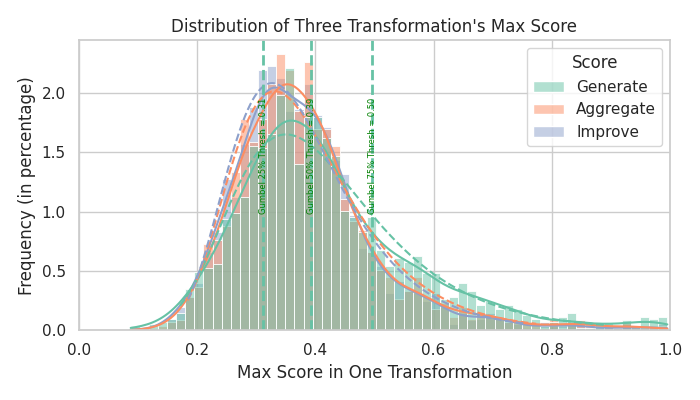} 
\caption{The distribution of the maximum score for each transformation. The solid line represents the kernel density estimation curve, while the dashed line represents the Gumbel distribution curve estimated according to the method in ~\ref{subsec:gumbelthreshold}. According to Equ.~\ref{GumbelThresh}, taking the generation transformation as an example, the corresponding scores for confidence levels of 25\%, 50\%, and 75\% are calculated.}
\label{fig4}
\end{center}
\end{figure}

As shown in Fig. \ref{fig3}, we plot the score distribution of training data under three transformations. For the corresponding transformation, the solid line is the kernel density estimation curve, while the dashed line is the mean score, where the generation transformation is slightly larger than the other two types.
Fig. \ref{fig4} shows the distribution of the maximum score for each transformation. Since our graph structure has three edges for each transformation, the maximum value is the highest score among the three answers. 
According to the method in Section~\ref{subsec:gumbelthreshold}, the Gumbel distribution calculated is represented by dashed lines. It can be observed that the Gumbel distribution can fit well with the distribution of maximum values for each transformation. In the reasoning process, the Gumbel threshold calculated from the distribution is used as the hyper-parameter of GoT.

\subsection{Main Results}
\label{sec:main-results}

\begin{figure}[b]
\begin{center}
\hspace{-0.0cm} 
\includegraphics[scale=0.4]{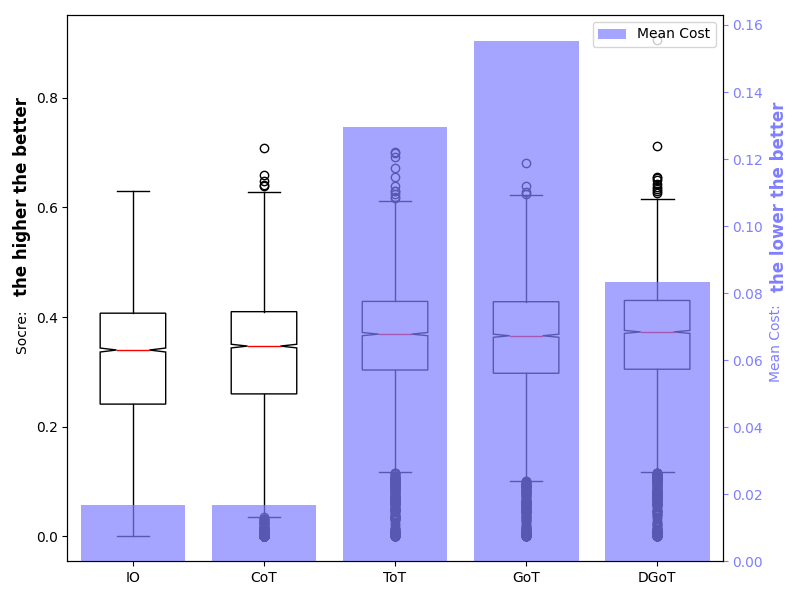} 
\caption{Scores and costs of abstract generated by ChatGLM2-6B under different prompt approaches.}
\label{fig:result}
\end{center}
\end{figure}

\begin{table*}[t]
\centering
\begin{tabular}{p{1cm}p{2cm}p{1cm}p{1cm}p{1.6cm}p{1.6cm}p{2.5cm}p{2cm}}
\hline
\textbf{Method} & \textbf{R-1} & \textbf{R-2} & \textbf{R-L} & \textbf{Prompt Tokens} & \textbf{Response Tokens} & \textbf{Cost} & \textbf{Cost-effectiveness}\\
\hline
IO & 0.303 & 0.081 & 0.166 & 10660.79 & \ \ 402.79 & 0.0167 \\
CoT & \textbf{0.314} & 0.083 & 0.171 & 10644.81 & \ \ 358.77 & \textbf{0.0166} \\
\hline
ToT & 0.356(0.042) & 0.098 & 0.190 & 82850.63 & 2606.48 & 0.1294 (0.1128) & 2.686\\
GoT & 0.354(0.040) & \textbf{0.099} & 0.190 & 99184.15 & 3219.40 & 0.1552 (0.1386) & 3.465\\
DGoT & \textbf{0.358}(0.044) & \textbf{0.099} & \textbf{0.192} & \textbf{53414.97} & \textbf{1565.12} & \textbf{0.0833} (0.0667) & \textbf{1.516}\\
\hline
\end{tabular}
\caption{Main Results. \textbf{R-1}, \textbf{R-2}, and \textbf{R-L} represent the ROUGE scores of the generated abstract and the actual abstract of the source articles respectively. \textbf{Prompt Tokens} is the average number of tokens input to LLM throughout the entire process of the method, while \textbf{Response Tokens} is the average number of tokens returned by LLM. \textbf{Cost} is the cost corresponding to the number of tokens. Here, we calculate the price of the local model based on that setting of chatgpt-3.5 (\$1.5/1M input tokens, \$2/1M output tokens).}
\label{mainresult}
\end{table*}

Table~\ref{mainresult} compares the ROUGE scores and costs of our method with other prompt approaches. The 5 approaches involved were validated on 5224 PubMedCite test data we processed, among which the threshold setting method for DGoT is Simple Mean Threshold. Since both IO and CoT only make one query to LLM, we compare the results of ToT, GoT, and DGoT with the best of the two methods. It can be seen that the multi-round query approach has significant performance improvement compared to single-round query.

In Figure~\ref{fig:result}, it can be visually observed that DGoT has lower costs compared to ToT or GoT. Additionally, DGoT demonstrates superior cost-effectiveness.

Cost-effectiveness is defined as the cost required to improve the performance of a unit metric compared to a baseline method. Therefore, the smaller the cost-effectiveness, the better. E.g. The values in the parentheses to the right of \textbf{cost} in Table~\ref{mainresult} (representing the additional cost required for multi-round query approaches compared to the best single-round query method CoT) divided by the values in the parentheses to the right of \textbf{R-1} (representing the improvement in R-1 score compared to CoT) yield the cost-effectiveness. Compared to fixed tree or graph structures, our method has significant cost advantages.

\subsection{Effects of threshold function $H$ setting}

\begin{figure}[b]
\begin{center}
\hspace{-0.0cm} 
\includegraphics[scale=0.4]{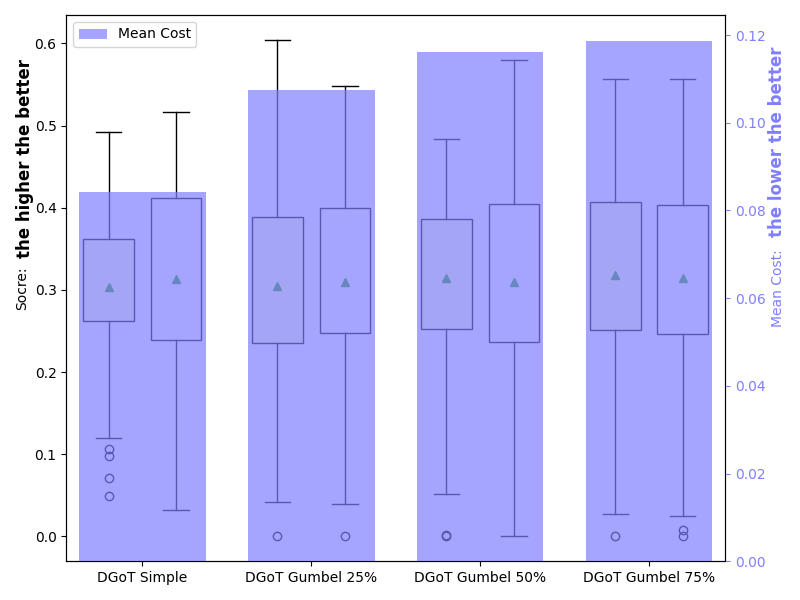} 
\caption{The effect of different threshold settings on output score and cost. Each associated with two box plots showing the ROUGE R-1 scores of the generated abstract compared to the original introduction and the actual abstract from left to right.}
\label{fig:gumbel}
\end{center}
\end{figure}

Table~\ref{gumbelTSet} compares the results under different threshold settings. This experiment was validated on 100 papers in the test set. Since the scores compared with the threshold are calculated in the generated abstract and introduction sections of the original document, as the threshold increases, the output result's Intro.R-1 score continues to increase. However, the Abst. ROUGE score compared with the actual abstract does not show a completely linear growth trend. 

Figure ~\ref{fig:gumbel} shows that as the threshold increases, the required costs continue to rise, but the rate of increase gradually decreases.

\begin{table}[t]
\centering
\begin{tabular}{p{1cm}p{0.8cm}p{0.8cm}p{0.8cm}p{0.8cm}p{1cm}}
\hline
\textbf{Setting} & \textbf{Intro. R-1} & \textbf{Abst. R-1} & \textbf{Abst. R-2} & \textbf{Abst. R-L} & \textbf{Cost}\\
\hline
Simple & 0.303 & 0.313 & 0.075 & \textbf{0.176} & \textbf{0.0841}\\
\hline
25\% & 0.305 & 0.309 & 0.070 & 0.173 & 0.1075\\
50\% & 0.314 & 0.309 & 0.075 & 0.170 & 0.1161\\
75\% & \textbf{0.317} & \textbf{0.314} & \textbf{0.078} & 0.175 & 0.1186\\
\hline
\end{tabular}
\caption{ROUGE scores and cost for different threshold settings. From top to bottom, it represents simple threshold settings, as well as Gumbel threshold settings with confidence levels of 25\%, 50\%, and 75\%.}
\label{gumbelTSet}
\end{table}

\section{Related Work} 

\subsection{Large Language Models}

In recent years, the scale of language models has been continuously increasing \cite{Devlin2019BERTPO, Radford2018ImprovingLU, Brown2020LanguageMA, OpenAI2023GPT4TR}, and they have also shown significant improvements in performance in various downstream tasks, known as emergent capabilities \cite{Wei2022EmergentAO}. The open-source large language model is becoming a new trend in the field of artificial intelligence. Recent work includes LLaMA \cite{Touvron2023LLaMAOA}, RWKV \cite{Peng2023RWKVRR}, OpenFlamingo \cite{Awadalla2023OpenFlamingoAO}, GLM-130B \cite{Zeng2022GLM130BAO} etc. Based on our method, the ceiling performance of locally deployed open-source models on the task of paper abstracting can be explored.

\subsection{Prompt Paradigms}
In section~\ref{sec:promptA}, we introduced some mainstream prompt methods \cite{Wei2022ChainOT, Yao2023TreeOT, Besta2023GraphOT}. From the perspective of graph theory, a path is a sequence of vertices in a graph, and a tree is an acyclic-connected graph. Therefore, under certain conditions, GoT can degenerate into ToT or CoT. Other prompt approaches include Self-consistent CoT \cite{Wang2022SelfConsistencyIC}, which samples a set of different reasoning paths and then selects the most consistent answer. 

Cumulative Reasoning \cite{Zhang2023CumulativeRW} decomposes the reasoning process into three components: proposer, verifier, and reporter, and uses an accumulative and iterative approach to reasoning. Chain-of-Verification \cite{Dhuliawala2023ChainofVerificationRH} drafts verification questions to promote model optimization of the initial output results. Due to the scalability of the GoT system architecture, the above methods can be incorporated into the system by adding functional modules. In this paper, the final graph form may be a path or a tree by dynamically adjusting the GoT graph structure, in order to minimize the reasoning cost while completing scientific paper abstract generation.

\subsection{LLM in Scientific Research}

Traditional AI-assisted scientific paper tasks typically involve training models on data of designated domains. The large language model can complete downstream tasks with few or zero-shot approaches, so its application in completing scientific research tasks has attracted increasing attention \cite{Birhane2023ScienceIT, Boiko2023EmergentAS, Bran2023ChemCrowAL, Liang2023CanLL}. Large language models trained on scientific paper datasets include Galactica \cite{Taylor2022GalacticaAL}, Darwin \cite{xie2023darwin} and Mozi \cite{Mozi2023} etc. This article has attempted to evaluate the performance of the universal open-source large model in the task of paper abstracts, and related evaluation tests can be conducted on the scientific article large model in the future.

\section{Conclusion}


In this work, we propose a Dynamic Graph of Thought prompt approach that can adaptively adjust the graph structure during the reasoning process to reduce the language model cost. We define a threshold-setting mechanism for the GoT evaluation function to provide a reference for the trade-off between performance and cost. Our experiments show that on the task of scientific literature abstract generation, this method achieves the best cost-effectiveness compared to other multi-round prompt approaches.

\section*{Acknowledgements}

This work was supported by National Demonstration Center for Experimental Electronic Information Education (Beijing University of Posts and Telecommunications) and computing resources are supported by the High-performance Computing Platform of BUPT.

\nocite{*}
\section*{Bibliographical References}\label{sec:reference}

\bibliographystyle{lrec-coling2024-natbib}
\bibliography{lrec-coling2024-example}


\clearpage 

\appendix
\section{Appendix}

\subsection{Other Potential Influencing Factors on the Results}

To investigate the effects of prompt length, Branching Factors, and different models on experimental outcomes, we conducted validations on the initial 100 records from both the training and testing sets of PubMedCite. These validations were conducted on a server equipped with a 24G 3090 graphics card. For ChatGLM2-6B, we deployed it using the same API approach as described in Section~\ref{sec:ImplementationDetails} of the main text. For InternLM2-Chat-7B\footnote{\href{https://github.com/InternLM/InternLM}{https://github.com/InternLM/InternLM}}, we utilized LMDeploy\footnote{\href{https://github.com/InternLM/lmdeploy}{https://github.com/InternLM/lmdeploy}} to accelerate its inference process. We also provided the inference duration for each method under each model for time comparison. The pertinent code is available in our GitHub repository\footnote{\href{https://github.com/JayceNing/DGoT}{https://github.com/JayceNing/DGoT}}, along with a readily deployable Docker image.

\subsubsection{Effect of Prompt Length}
\label{sec:EffectofPromptLength}

\begin{figure}[!bp]
\begin{center}
\hspace{-0.0cm} 
\includegraphics[scale=0.4]{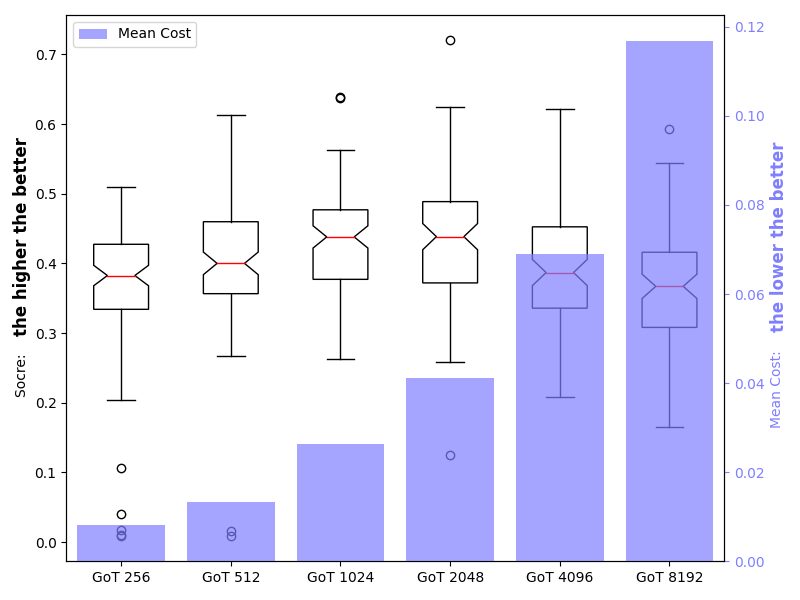} 
\caption{Scores and costs of abstract generated by ChatGLM2 using GoT prompt approach under various prompt length settings.}
\label{fig:PromptLengthChatGLM2}
\end{center}
\end{figure}

\begin{figure}[!htp]
\begin{center}
\hspace{-0.0cm} 
\includegraphics[scale=0.4]{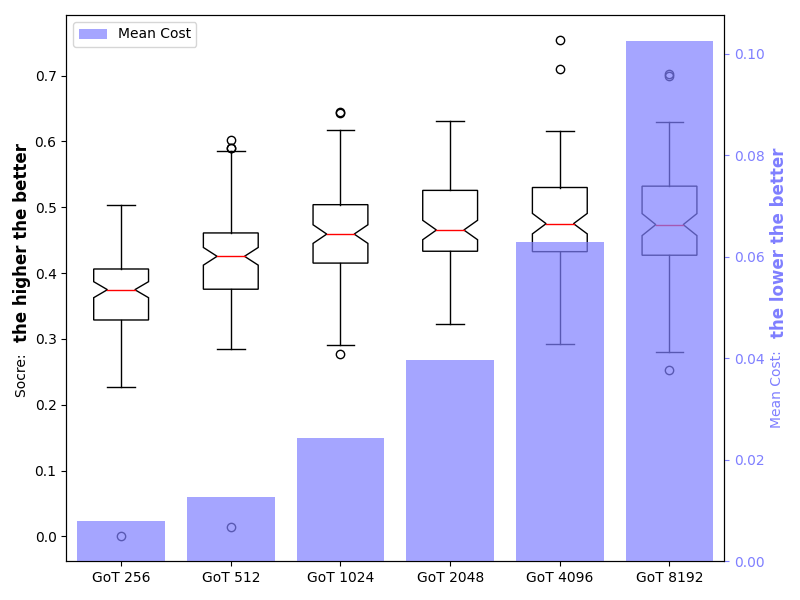} 
\caption{Scores and costs of abstract generated by InternLM2-chat-7B using GoT prompt approach under various prompt length settings.}
\label{fig:PromptLengthInternLM2}
\end{center}
\end{figure}

In Section~\ref{sec:ImplementationDetails}, longer input texts are truncated before inputting the model. In order to explore the influence of prompt length (Includes prompt text within the prompt framework, along with the corresponding filled-in information) on the result, we adopt the same GoT setting as Section~\ref{sec:Baselines} to validate the two models.

For prompt length's effect on R-1 scores, Figure~\ref{fig:PromptLengthChatGLM2} highlights ChatGLM2's peak R-1 score at input length 2048, while Figure~\ref{fig:PromptLengthInternLM2} shows InternLM2's peak at length 4096.  Both models exhibit an initial increase followed by a decrease in R-1 scores as prompt length increases.

Table~\ref{EffectofPromptLength} lists the same result metrics as in Section~\ref{sec:main-results}, in addition to the proportion of truncated input text and the inference time for each prompt length. Longer input leads to higher inference time and cost. The rate of input truncation is not linearly correlated with the resulting scores. Different models exhibit varied inference performance, with ChatGLM2 excelling at retaining information from the introduction, while InternLM2 demonstrates stronger abstract generation capabilities.

\subsubsection{Effect of Branching Factors}
\label{sec:EffectofBranchingFactors}

In ~\ref{sec:EffectofPromptLength}, InternLM2 demonstrates its best abstract generation capability when the prompt length is 4096. Under this condition, we test the impact of different branching factors $k$ on the results.

Figure~\ref{fig:BranchingFactorsInternLM2} demonstrates that as the branching factor $k$ increases, the R-1 scores between the output and the original introduction gradually improve. This is because the scoring function is designed to select responses with the highest R-1 scores compared to the introduction. However, there is no improvement in R-1 scores between the output and the actual abstract.

Table~\ref{tab:EffectofBranchingFactors} shows that the Introduction R-1 scores gradually increase as the branching factor grows, but the rate of increase diminishes. This is reflected in the rise of Cost-effectiveness, indicating that the cost required to improve each unit score becomes increasingly higher. This to some extent demonstrates that the number of Agents also adheres to Scaling Laws \cite{Kaplan2020ScalingLF}, but the performance gain it brings is not limitless.

\begin{figure*}[!htp]
\begin{center}
\hspace{-1.3cm} 
\includegraphics[scale=0.15]{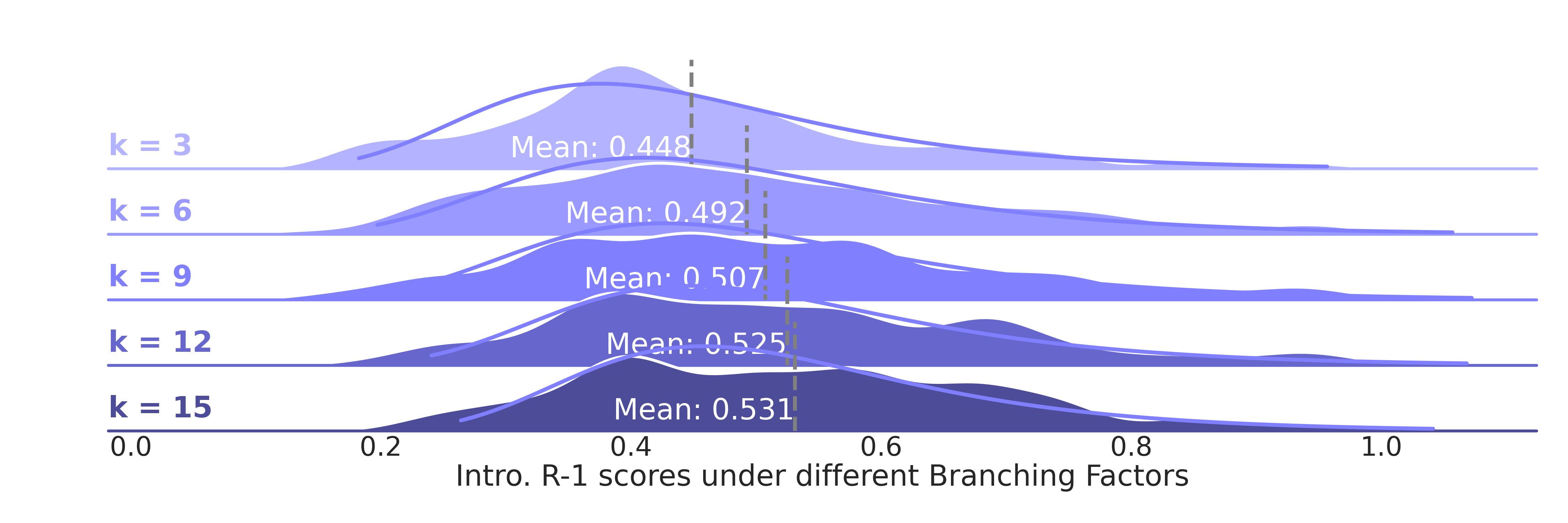} 
\caption{The effect of different branching factors $k$ settings on output Intro. R-1 score. }
\label{fig:NodesNum}
\end{center}
\end{figure*}

\begin{figure}[!htp]
\begin{center}
\hspace{-0.0cm} 
\includegraphics[scale=0.4]{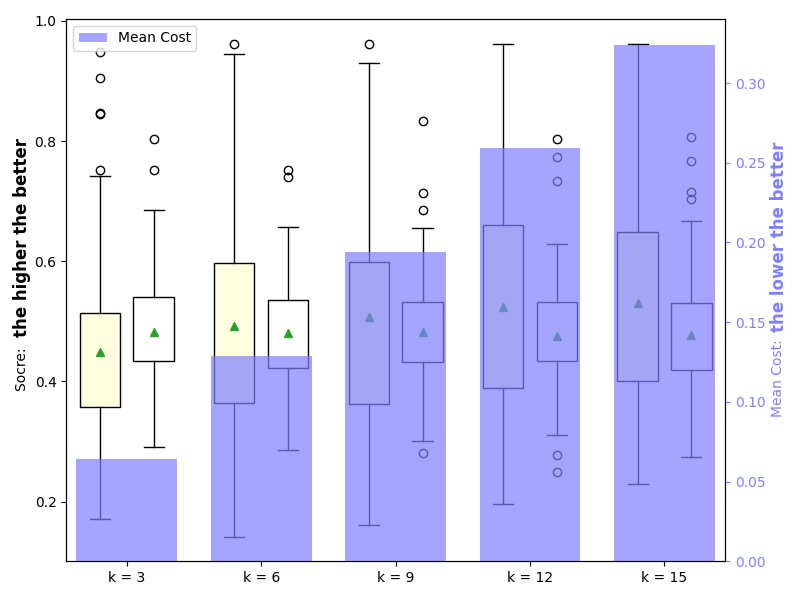} 
\caption{The effect of different branching factors $k$ settings on output score and cost. Each $k$ setting corresponds to two box plots, representing the ROUGE R-1 scores of the generated abstract compared to the original literature introduction and the actual abstract from left to right.}
\label{fig:BranchingFactorsInternLM2}
\end{center}
\end{figure}

Figure~\ref{fig:NodesNum} provides a more intuitive presentation of this conclusion. Its horizontal axis represents the

\begin{table*}[!htp]
\centering
\begin{tabular}{p{1.09cm}p{1cm}p{0.9cm}p{0.9cm}p{0.9cm}p{0.9cm}p{0.9cm}p{1.41cm}p{1.3cm}p{1.01cm}p{0.8cm}}
\hline
\textbf{Model/ Method} & \textbf{Prompt Length} & \textbf{Cut Ratio} & \textbf{Intro. R-1} & \textbf{Abst. R-1} & \textbf{Abst. R-2} & \textbf{Abst. R-L} & \textbf{Prompt Tokens} & \textbf{Resp. Tokens} & \textbf{Infer. Time(s)} & \textbf{Cost}\\
\hline
Chat- & \hspace{0.08cm} 256 & 0.993 & 0.339 & 0.367 & 0.091 & 0.189 & \hspace{0.08cm} 2623.29 & 2062.74 & \hspace{0.08cm} 76.61 & 0.008 \\
GLM2/ & \hspace{0.08cm} 512 & 0.973 & 0.456 & 0.400 & 0.111 & 0.192 & \hspace{0.08cm} 5360.43 & 2613.06 & \hspace{0.08cm} 96.05 & 0.013 \\
GoT & 1024 & 0.852 & 0.517 & 0.431 & 0.144 & 0.216 & 12881.78 & 3545.09 & 126.69 & 0.026 \\
& 2048 & 0.775 & \textbf{0.520} & 0.435 & 0.154 & 0.221 & 23189.33 & 3223.30 & 120.00 & 0.041 \\
& 4096 & 0.693 & 0.510 & 0.391 & 0.132 & 0.203 & 40675.92 & 3979.97 & 154.09 & 0.068 \\
& 8192 & 0.465 & 0.436 & 0.366 & 0.104 & 0.191 & 70249.50 & 5711.58 & 282.49 & 0.116 \\
\hline
Intern- & \hspace{0.08cm} 256 & 0.993 & 0.317 & 0.368 & 0.097 & 0.187 & \hspace{0.08cm} 3368.73 & 1440.99 & \hspace{0.08cm} 37.16 & 0.007 \\
LM2/ & \hspace{0.08cm} 512 & 0.973 & 0.450 & 0.418 & 0.125 & 0.200 & \hspace{0.08cm} 5949.60 & 1892.68 & \hspace{0.08cm} 47.75 & 0.012 \\
GoT & 1024 & 0.830 & 0.418 & 0.456 & 0.164 & 0.235 & 13642.57 & 1894.43 & \hspace{0.08cm} 48.89 & 0.024 \\
& 2048 & 0.740 & 0.447 & 0.471 & 0.176 & 0.240 & 23849.84 & 1965.58 & \hspace{0.08cm} 53.24 & 0.039 \\
& 4096 & 0.670 & 0.447 & \textbf{0.482} & \textbf{0.190} & \textbf{0.259} & 39069.64 & 2139.46 & \hspace{0.08cm} 60.76 & 0.062 \\
& 8192 & 0.436 & 0.422 & 0.479 & 0.183 & 0.250 & 65586.77 & 2049.10 & \hspace{0.08cm} 72.41 & 0.102 \\
\hline
\end{tabular}
\caption{Effect of Prompt Length. The performance of ChatGLM2 and InternLM2 models using GoT as prompt approach was tested under different input prompt lengths. If the length of the input exceeds the specified \textbf{Prompt Length}, the input will be truncated. The \textbf{Cut Ratio} indicates the proportion of the input to be truncated. \textbf{Intro. R-1} represents the ROUGE scores of the generated abstract and the introduction of the source articles. \textbf{Abst. R-1}, \textbf{Abst. R-2}, and \textbf{Abst. R-L} represent the ROUGE scores of the generated abstract and the actual abstract of the source articles respectively. \textbf{Prompt Tokens} is the average number of tokens input to LLM throughout the entire process of the method, while \textbf{Resp. Tokens} is the average number of tokens returned by LLM. \textbf{Infer. Time} refers to the time required for a specific model to generate the abstract of a research paper using GoT method. \textbf{Cost} is the cost corresponding to the number of tokens. Here, we calculate the price of the local model based on that setting of chatgpt-3.5 (\$1.5/1M input tokens, \$2/1M output tokens).}
\label{EffectofPromptLength}
\end{table*}

\begin{table*}[!htp]
\centering
\begin{tabular}{p{0.4cm}p{2cm}p{0.9cm}p{0.9cm}p{0.9cm}p{1.59cm}p{1.4cm}p{1.02cm}p{1.8cm}p{1cm}}
\hline
\hspace{0.08cm} $k$ & \textbf{Introduction R-1} & \textbf{Abst. R-1} & \textbf{Abst. R-2} & \textbf{Abst. R-L} & \textbf{Prompt Tokens} & \textbf{Resp. Tokens} & \textbf{Infer. Time(s)} & \textbf{Cost} & \textbf{C/E}\\
\hline
\hspace{0.08cm} 3 & 0.448 & \textbf{0.481} & \textbf{0.191} & \textbf{0.264} & \hspace{0.08cm} 39940.46 & \hspace{0.08cm} 2088.17 & \hspace{0.08cm} 60.86 & 0.064 & \\
\hline
\hspace{0.08cm} 6 & 0.492(0.044) & 0.480 & 0.187 & 0.250 & \hspace{0.08cm} 80091.36 & \hspace{0.08cm} 4360.62 & 122.64 & 0.128(0.064) & \textbf{1.454}\\
\hspace{0.08cm} 9 & 0.507(0.059) & 0.481 & 0.188 & 0.255 & 120196.65 & \hspace{0.08cm} 6761.61 & 187.29 & 0.193(0.129) & 2.186\\
12 & 0.524(0.076) & 0.475 & 0.190 & 0.255 & 160543.10 & \hspace{0.08cm} 9139.17 & 251.58 & 0.259(0.195) & 2.565\\
15 & \textbf{0.530}(0.082) & 0.478 &  0.187 & 0.251 & 200225.20 & 11821.69 & 322.74 & 0.323(0.259) & 3.158\\

\hline
\end{tabular}
\caption{Effect of Branching Factors $k$. C/E represents Cost-effectiveness (see section~\ref{sec:main-results}).}
\label{tab:EffectofBranchingFactors}
\end{table*}

\noindent Intro. R-1 scores. The purple line shows the Gumbel distribution curve, and the gray dashed line marks the mean position. As the branching factor rises, the increase in average output score diminishes.

\subsubsection{Results under Optimal Prompt Length}

\begin{table*}[!htp]
\hspace{-0.0cm}
\begin{tabular}{p{1.02cm}p{0.64cm}p{1.7cm}p{0.7cm}p{0.7cm}p{0.7cm}p{1.44cm}p{1.24cm}p{0.9cm}p{1.7cm}p{0.8cm}}
\hline
\textbf{Model} & \textbf{Meth- od} & \textbf{Introduction R-1} & \textbf{Abst. R-1} & \textbf{Abst. R-2} & \textbf{Abst. R-L} & \textbf{Prompt Tokens} & \textbf{Resp. Tokens} & \textbf{Infer. Time(s)} & \textbf{Cost} & \textbf{C/E}\\
\hline
Chat- & IO & \textbf{0.311} & 0.387 & 0.126 & 0.204 & \hspace{0.16cm}2274.14 & \hspace{0.16cm}233.51 & 10.53 & 0.003 & \\
GLM2 & CoT & 0.305 & \textbf{0.401} & \textbf{0.129} & \textbf{0.213} & \hspace{0.16cm}2269.77 & \hspace{0.16cm}214.12 & \hspace{0.16cm}9.84 & 0.003 &  \\
\hline
 & ToT & \textbf{0.476}(0.171) & 0.390 & \textbf{0.130} & \textbf{0.199} & 20465.34 & 2376.15 & 96.11 & 0.035(0.032) & 0.187 \\
 & GoT & 0.475(0.170) & 0.382 & 0.128 & 0.196 & 20409.60 & 2442.31 & 97.25 & 0.035(0.032) & 0.188 \\
 & DGoT & 0.418(0.113) & \textbf{0.395} &  0.129 & \textbf{0.199} & 10602.23 & 1256.39 & 55.19 & 0.018(0.015) & \textbf{0.132} \\
 \hline
Intern- & IO & \textbf{0.317} & \textbf{0.439} & \textbf{0.164} & \textbf{0.242} & \hspace{0.16cm}4420.49 & \hspace{0.16cm}239.16 & \hspace{0.16cm}8.14 & 0.007 & \\
LM2 & CoT & 0.279 & 0.436 & 0.158 & 0.237 & \hspace{0.16cm}4417.75 & \hspace{0.16cm}195.71 & \hspace{0.16cm}7.19 & 0.007 &  \\
\hline
 & ToT & \textbf{0.477}(0.198) & 0.414 & 0.148 & 0.212 & 39812.07 & 2241.35 & 67.43 & 0.064(0.057) & 0.287 \\
 & GoT & 0.456(0.177) & 0.419 & \textbf{0.156} & 0.220 & 39732.42 & 2225.52 & 67.26 & 0.064(0.057) & 0.322  \\
 & DGoT & 0.399(0.120) & \textbf{0.422} &  0.152 & \textbf{0.222} & 19690.67 & 1016.34 & 33.78 & 0.031(0.024) & \textbf{0.200} \\

\hline
\end{tabular}
\caption{Results under Optimal Prompt Length.}
\label{tab:AddMainResults}
\end{table*}

Section~\ref{sec:EffectofPromptLength} Experimental results show that the optimal prompt length for the ChatGLM2 model is 2048, while for InternLM2 it is 4096. Section~\ref{sec:EffectofBranchingFactors} demonstrates that the optimal branching factor $k$ is 3. Under this experimental configuration, the performance of both models is tested using different prompt approaches. The dataset used for testing consists of the first 100 entries in the test set, with DGoT threshold set to Simple Mean Threshold.

\begin{figure}[!tp]
\begin{center}
\hspace{-0.0cm} 
\includegraphics[scale=0.4]{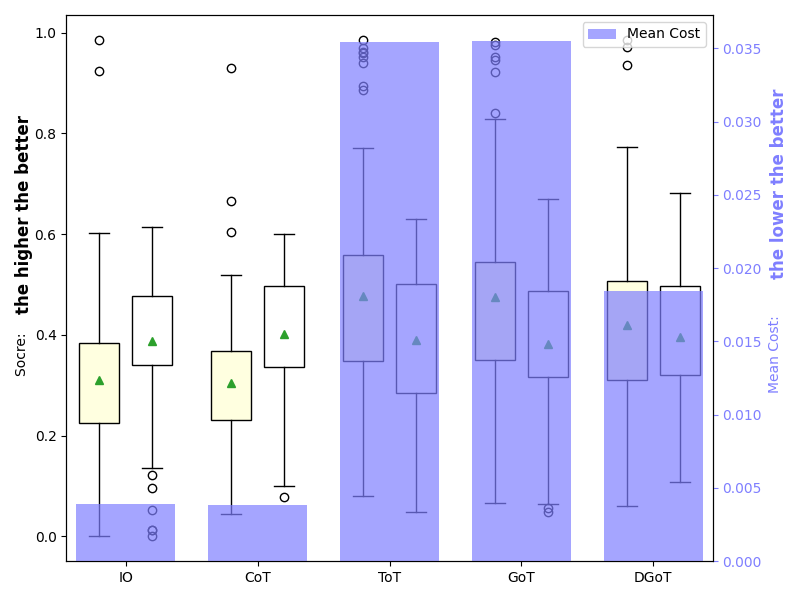} 
\caption{Scores and costs of abstract generated by ChatGLM2-6B under different prompt approaches.}
\label{fig:AddMainResultsChatGLM2}
\end{center}
\end{figure}

\begin{figure}[!tp]
\begin{center}
\hspace{-0.0cm} 
\includegraphics[scale=0.4]{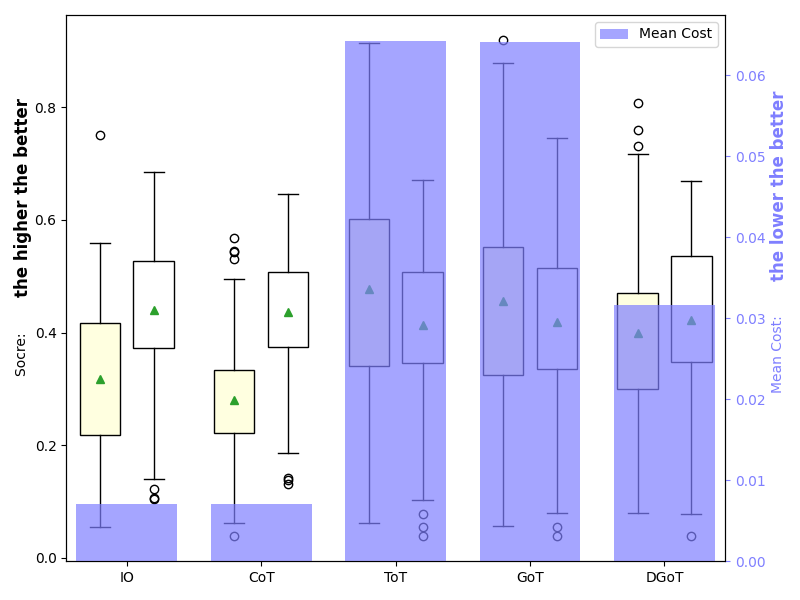} 
\caption{Scores and costs of abstract generated by InternLM2-Chat-7B under different prompt approaches.}
\label{fig:AddMainResultsInternLM2}
\end{center}
\end{figure}

Figure~\ref{fig:AddMainResultsChatGLM2} and~\ref{fig:AddMainResultsInternLM2} present the experimental results of the two models on a small test set, which roughly align with the trends discussed in Section~\ref{sec:main-results}. Additionally, here we provide the R-1 scores of the output results compared to the original text introductions. It can be observed that, compared to single-round query method, multi-round query approach generally shows improvement in this score.

Table~\ref{tab:AddMainResults} presents more detailed results. The highest ROUGE scores obtained from different prompting approaches within the two major categories of single-round query and multi-round query are highlighted in bold. For different models, using IO or CoT for single-round query yields varying results. InternLM2 performs better with IO, while ChatGLM2 performs better with CoT. DGoT, as a multi-round query approach, generally demonstrates good performance in ROUGE scores for generating abstract compared to the original abstract, with superior Cost-effectiveness.

Nevertheless, the performance of Intro. R-1 of ToT is better than that of GoT in the current multi-round query approach, which indicates that aggregation transformation under the current prompt word framework does not bring performance improvement compared with generation transformation. In addition, the increase in Intro. R-1 scores was accompanied by a decrease in Abst. R-1 scores. This indicates that there is a tradeoff between these two evaluation indicators. Finding a suitable evaluation function to improve the output of the prompt method is still a problem worth studying.

\subsection{Prompt Framework for Transformations}
\label{sec:OtherPromptFrameworks}
\begin{figure*}[ ]
\begin{center}
\includegraphics[scale=0.53]{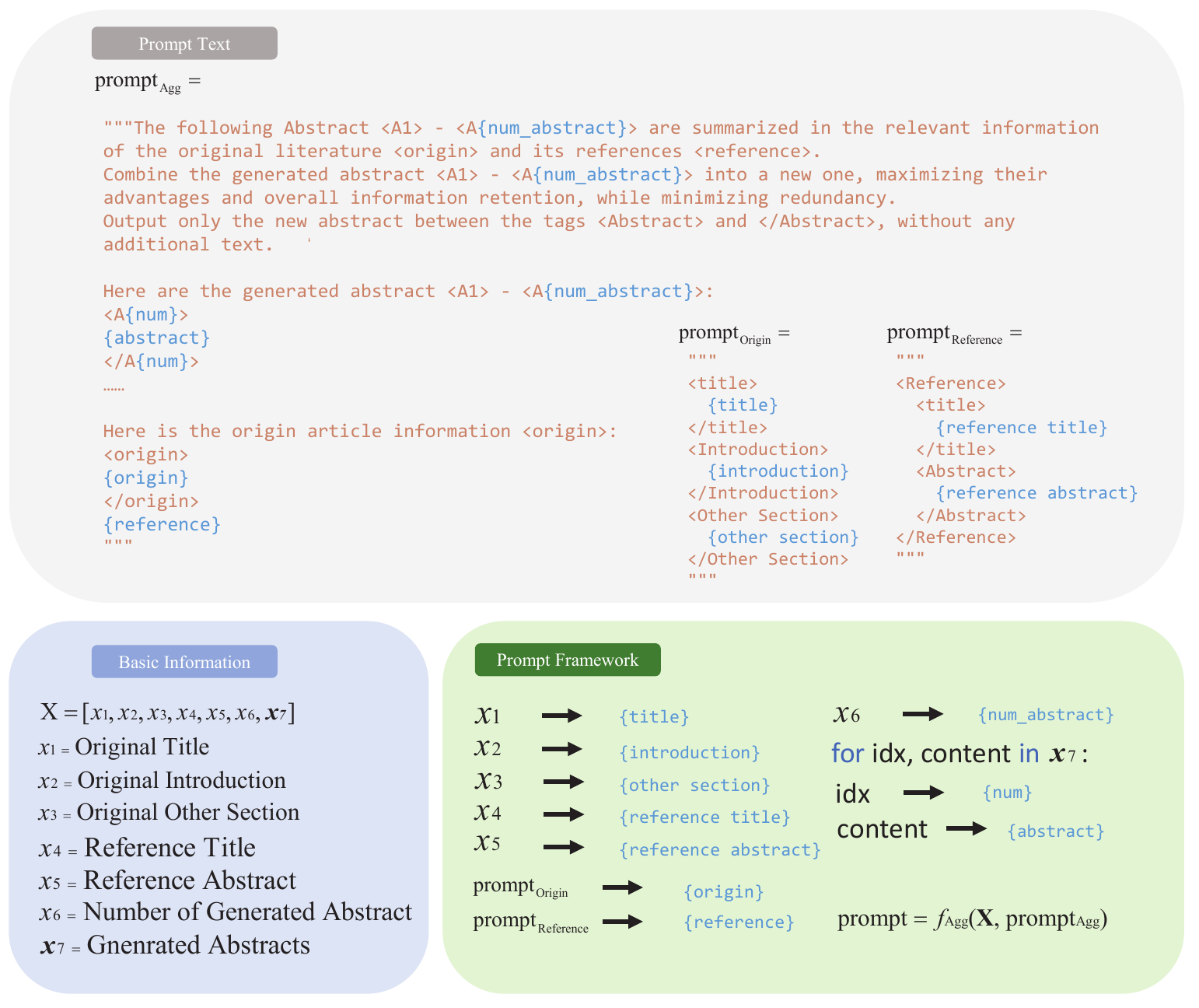} 
\caption{Prompt framework for Aggregation Transformation.}
\label{fig:AggPrompt}
\end{center}
\end{figure*}

\begin{figure*}[ ]
\begin{center}
\includegraphics[scale=0.53]{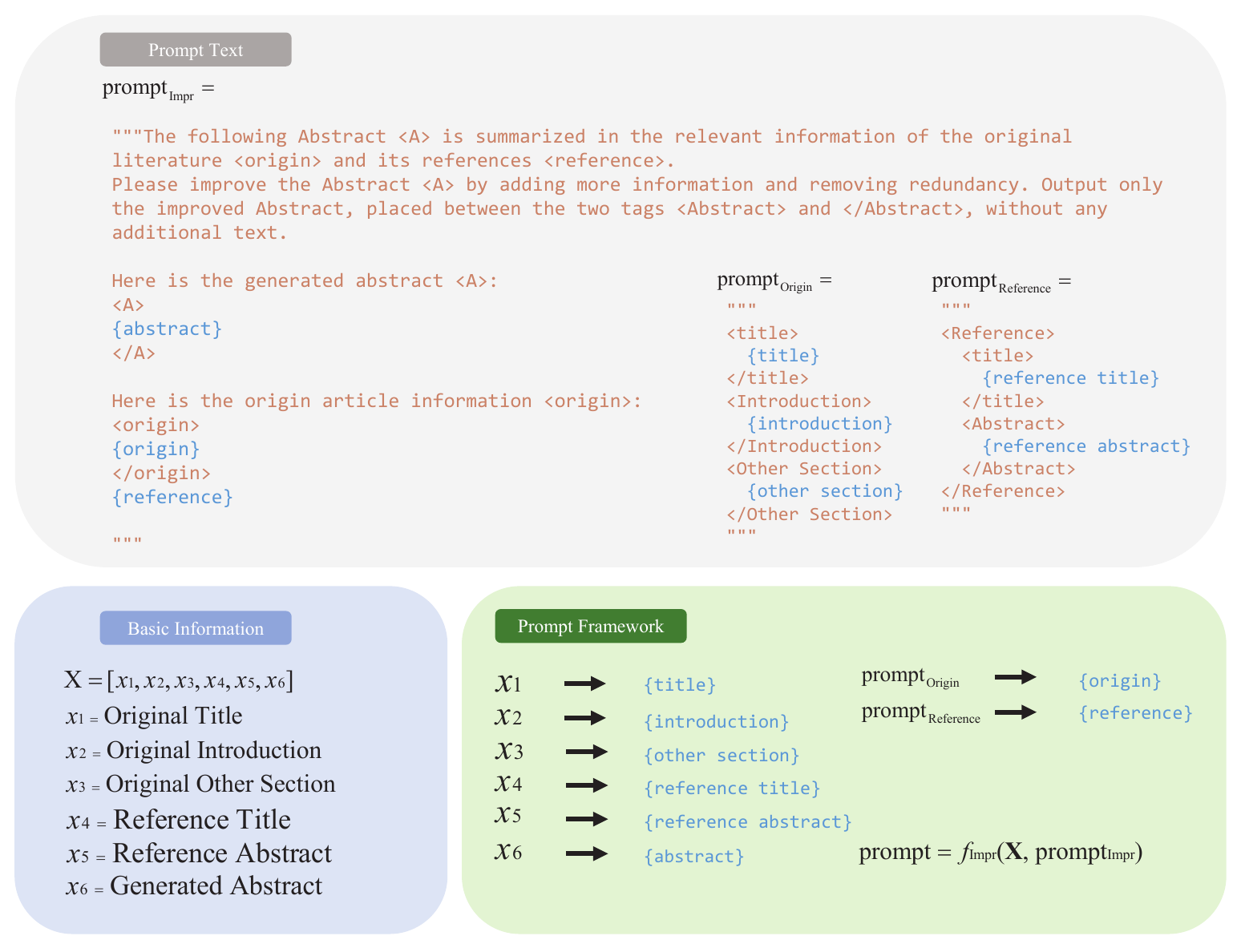} 
\caption{Prompt framework for Improving Transformation.}
\label{fig:ImprPrompt}
\end{center}
\end{figure*}

In the main text, Figure~\ref{fig:prompt} shows the prompt framework for generation transformation. Here, Figures~\ref{fig:AggPrompt} and~\ref{fig:ImprPrompt} show the prompt frameworks for aggregation transformation and improving transformation, respectively.

\end{document}